\documentclass{article} 
\usepackage{MSRA_TR,times}
\usepackage{hyperref}
\usepackage{url}
\usepackage{latexsym}
\usepackage{amsmath}
\usepackage{graphicx} 
\usepackage{multirow}
\usepackage{multicol}
\usepackage{algorithm}    
\usepackage{algpseudocode}
\usepackage{amssymb}
\usepackage{ucs}
\usepackage{enumitem}
\usepackage{color}

\title{Reaching Human-level Performance in \\ Automatic Grammatical Error Correction: \\  An Empirical Study}

\author{Tao Ge, Furu Wei, Ming Zhou \\
Natural Language Computing Group, Microsoft Research Asia, Beijing, China\\
\texttt{\{tage, fuwei, mingzhou\}@microsoft.com}
}

\iclrfinalcopy 

\begin{document}

\maketitle

\begin{abstract}
Neural sequence-to-sequence (seq2seq) approaches have proven to be successful in grammatical error correction (GEC). Based on the seq2seq framework, we propose a novel fluency boost learning and inference mechanism. Fluency boosting learning generates diverse error-corrected sentence pairs during training, enabling the error correction model to learn how to improve a sentence's fluency from more instances, while fluency boosting inference allows the model to correct a sentence incrementally with multiple inference steps. Combining fluency boost learning and inference with convolutional seq2seq models, our approach achieves the state-of-the-art performance: 75.72 ($F_{0.5}$) on CoNLL-2014 10 annotation dataset and 62.42 (GLEU) on JFLEG test set respectively, becoming the first GEC system that reaches human-level performance (72.58 for CoNLL and 62.37 for JFLEG) on both of the benchmarks.
\end{abstract}

\section{Introduction}\label{sec:intro}

\begin{figure}[h]
\centering
\includegraphics[width=9cm]{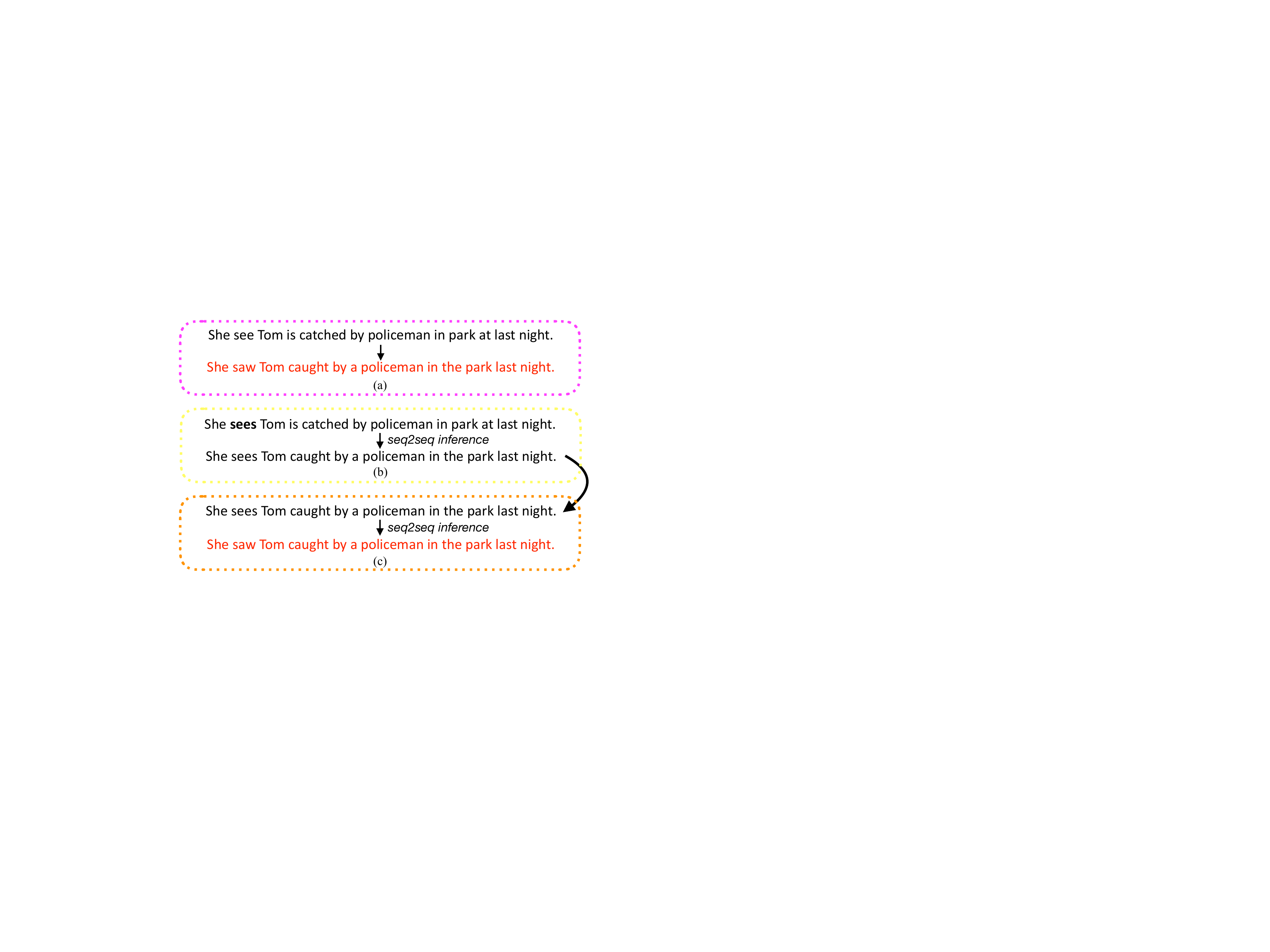}\vspace{-0.2cm}
\caption{\textbf{(a)} an error-corrected sentence pair; \textbf{(b)} if the sentence becomes slightly different, the model fails to correct it perfectly; \textbf{(c)} single-round seq2seq inference cannot perfectly correct the sentence, but multi-round inference can.}\label{fig:limitations}\vspace{-0.1cm}
\end{figure}

Sequence-to-sequence (seq2seq) models \citep{cho-EtAl:2014:EMNLP2014,DBLP:journals/corr/SutskeverVL14} for grammatical error correction (GEC) have drawn growing attention  \citep{yuan2016grammatical,xie2016neural,ji2017nested,schmaltz-EtAl:2017:EMNLP2017,sakaguchi2017grammatical,chollampatt2018,junczys2018approaching} in recent years. However, most of the seq2seq models for GEC have two flaws. \textbf{First}, the seq2seq models are trained with only limited error-corrected sentence pairs like Figure \ref{fig:limitations}(a). Limited by the size of training data, the models with millions of parameters may not be well generalized. Thus, it is common that the models fail to correct a sentence perfectly even if the sentence is slightly different from the training instance, as illustrated by Figure \ref{fig:limitations}(b). \textbf{Second}, the seq2seq models usually cannot perfectly correct a sentence with many grammatical errors through single-round seq2seq inference, as shown in Figure \ref{fig:limitations}(b) and \ref{fig:limitations}(c), because some errors in a sentence may make the context strange, which confuses the models to correct other errors. 

\begin{figure*}[t]
\centering
\includegraphics[width=14cm]{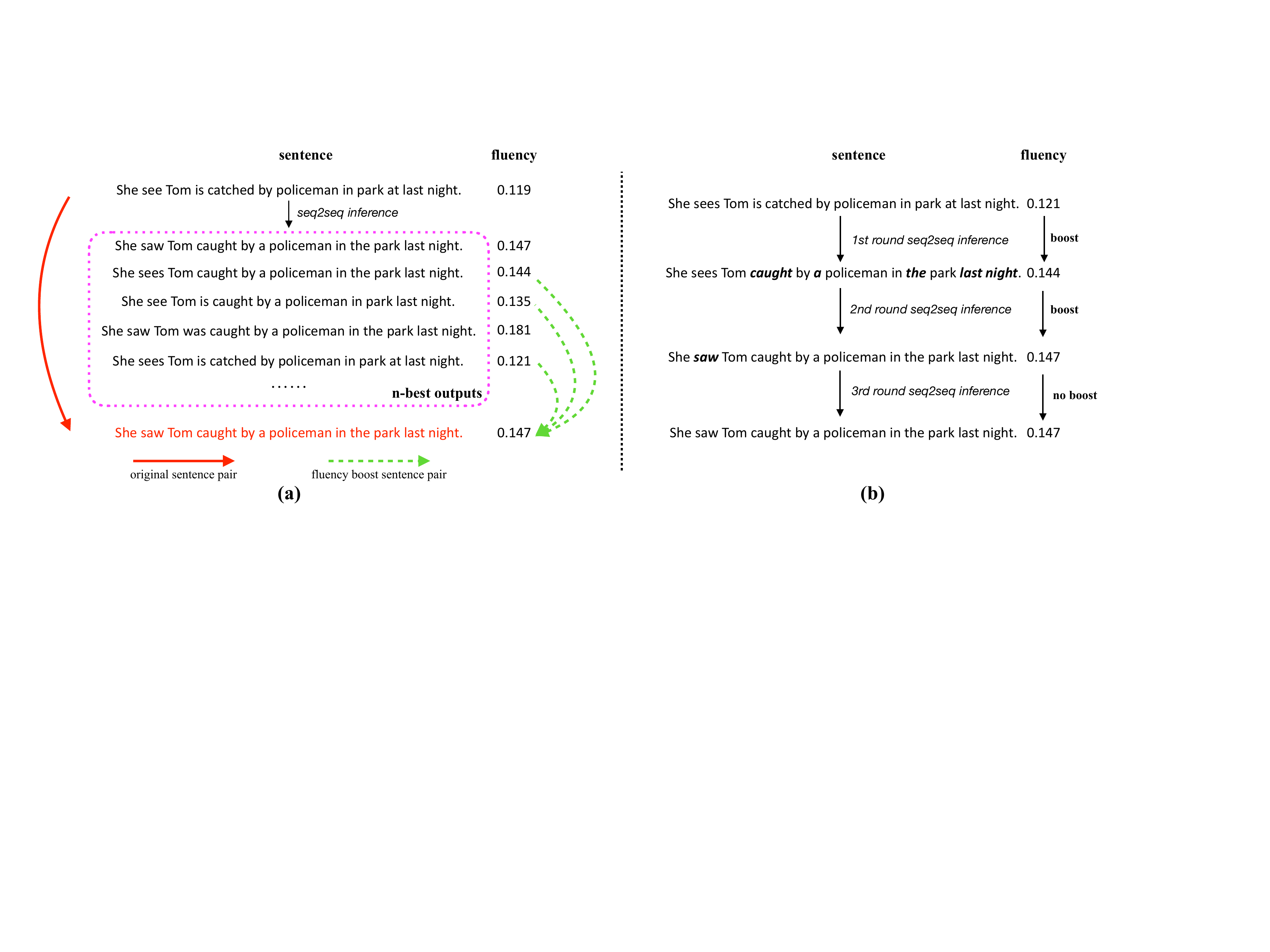}\vspace{-0.5cm}
\caption{Fluency boost learning and inference: \textbf{(a)} given a training instance (i.e., an error-corrected sentence pair), fluency boost learning establishes multiple fluency boost sentence pairs from the seq2seq's n-best outputs during training. The fluency boost sentence pairs will be used as training instances in subsequent training epochs, which helps expand the training set and accordingly benefits model learning; \textbf{(b)} fluency boost inference allows an error correction model to correct a sentence incrementally through multi-round seq2seq inference as long as its fluency can be improved.}\label{fig:fb}
\end{figure*}

To address the above-mentioned limitations in model learning and inference, we propose a novel fluency boost learning and inference mechanism, illustrated in Figure \ref{fig:fb}.

For fluency boosting learning, not only is a seq2seq model trained with original error-corrected sentence pairs, but also it generates less fluent sentences (e.g., from its n-best outputs) to establish new error-corrected sentence pairs by pairing them with their correct sentences during training, as long as the sentences' fluency\footnote{A sentence's fluency score is defined to be inversely proportional to the sentence's cross entropy, as is in Eq (\ref{eq:fluency}).} is below that of their correct sentences, as Figure \ref{fig:fb}(a) shows. Specifically, we call the generated error-corrected sentence pairs \textbf{fluency boost sentence pairs} because the sentence in the target side always improves fluency over that in the source side. The generated fluency boost sentence pairs during training will be used as additional training instances during subsequent training epochs, allowing the error correction model to see more grammatically incorrect sentences during training and accordingly improving its generalization ability. 



For model inference, fluency boost inference mechanism allows the model to correct a sentence incrementally with multi-round inference as long as the proposed edits can boost the sentence's fluency, as Figure \ref{fig:fb}(b) shows. For a sentence with multiple grammatical errors, some of the errors will be corrected first. The corrected parts will make the context clearer, which may benefit the model to correct the remaining errors. Moreover, based on the special characteristics of this task that the output prediction can be repeatedly edited and the basic fluency boost inference idea, we further propose a round-way correction approach that uses two seq2seq models whose decoding orders are left-to-right and right-to-left respectively. For round-way correction, a sentence will be corrected successively by the right-to-left and left-to-right seq2seq model\footnote{For convenience, we call the seq2seq model with right-to-left decoder right-to-left seq2seq model and the seq2seq model with left-to-right decoder left-to-right seq2seq model.}. Since the left-to-right and right-to-left decoder decode a sequence with different contexts, they have their unique advantages for specific error types. Round-way correction can fully exploit their pros and make them complement each other, which results in a significant improvement of recall.

Experiments show that combining fluency boost learning and inference with convolutional seq2seq models, our best GEC system\footnote{Our systems' outputs for CoNLL-2014 and JFLEG test set are available at \url{https://github.com/getao/human-performance-gec}} achieves 75.72 $F_{0.5}$ on CoNLL-2014 10 annotation dataset and 62.42 $GLEU$ on JFLEG test set, becoming the first system reaching human-level performance on both of the GEC benchmarks.

\section{Background: Neural grammatical error correction}\label{sec:background}

As neural machine translation (NMT), a typical neural GEC approach uses an encoder-decoder seq2seq model \citep{DBLP:journals/corr/SutskeverVL14,cho-EtAl:2014:EMNLP2014} with attention mechanism \citep{DBLP:journals/corr/BahdanauCB14} to edit a raw sentence into the grammatically correct sentence it should be, as Figure \ref{fig:limitations}(a) shows.

Given a raw sentence $\boldsymbol{x^r}=(x^r_1,\cdots,x^r_M)$ and its corrected sentence $\boldsymbol{x^c}=(x^c_1,\cdots,x^c_N)$ in which $x^r_M$ and $x^c_N$ are the $M$-th and $N$-th words of sentence $\boldsymbol{x^r}$ and $\boldsymbol{x^c}$ respectively, the error correction seq2seq model learns a probabilistic mapping $P(\boldsymbol{x^c}|\boldsymbol{x^r})$ from error-corrected sentence pairs through maximum likelihood estimation (MLE), which learns model parameters $\boldsymbol{\Theta_{crt}}$ to maximize the following equation:

\begin{equation}
\boldsymbol{\Theta^*_{crt}} = \arg \max_{\boldsymbol{\Theta_{crt}}} \sum_{(\boldsymbol{x^r},\boldsymbol{x^c}) \in \mathcal{S}^*} \log P(\boldsymbol{x^c}|\boldsymbol{x^r};\boldsymbol{\Theta_{crt}})
\end{equation}
where $\mathcal{S}^*$ denotes the set of error-corrected sentence pairs.



For model inference, an output sequence $\boldsymbol{x^o}=(x^o_1,\cdots,x^o_i,\cdots,x^o_L)$ is selected through beam search, which maximizes the following equation:

%
%
%

%
%
%
%
%
%
%

\begin{equation}
P(\boldsymbol{x^o}|\boldsymbol{x^r}) = \prod_{i=1}^{L}P(x^o_i|\boldsymbol{x^r},\boldsymbol{x^o}_{<i};\boldsymbol{\Theta_{crt}})
\end{equation}


\section{Fluency boost learning}\label{sec:learning}

Conventional seq2seq models for GEC learn model parameters only from original error-corrected sentence pairs.
However, such error-corrected sentence pairs are not sufficiently available. As a result, many neural GEC models are not very well generalized.

Fortunately, neural GEC is different from NMT. For neural GEC, its goal is improving a sentence's fluency\footnote{Fluency of a sentence in this work refers to how likely the sentence is written by a native speaker. In other words, if a sentence is very likely to be written by a native speaker, it should be regarded highly fluent.} without changing its original meaning; thus, any sentence pair that satisfies this condition (we call it \textbf{fluency boost condition}) can be used as a training instance.

%
%
%

In this work, we define $f(\boldsymbol{x})$ as the fluency score of a sentence $\boldsymbol{x}$:
\vspace{-0.05cm}
\begin{equation}\label{eq:fluency}
f(\boldsymbol{x})=\frac{1}{1+H(\boldsymbol{x})}
\end{equation}

\begin{equation}
H(\boldsymbol{x})=-\frac{\sum_{i=1}^{|\boldsymbol{x}|}\log P(x_i|\boldsymbol{x}_{<i})}{|\boldsymbol{x}|}
\end{equation}

\noindent where $P(x_i|\boldsymbol{x}_{<i})$ is the probability of $x_i$ given context $\boldsymbol{x}_{<i}$, computed by a language model, and $|\boldsymbol{x}|$ is the length of sentence $\boldsymbol{x}$. $H(\boldsymbol{x})$ is actually the cross entropy of the sentence $\boldsymbol{x}$, whose range is $[0,+\infty)$. Accordingly, the range of $f(\boldsymbol{x})$ is $(0,1]$.

\begin{figure}[t]
\centering
\includegraphics[width=13cm]{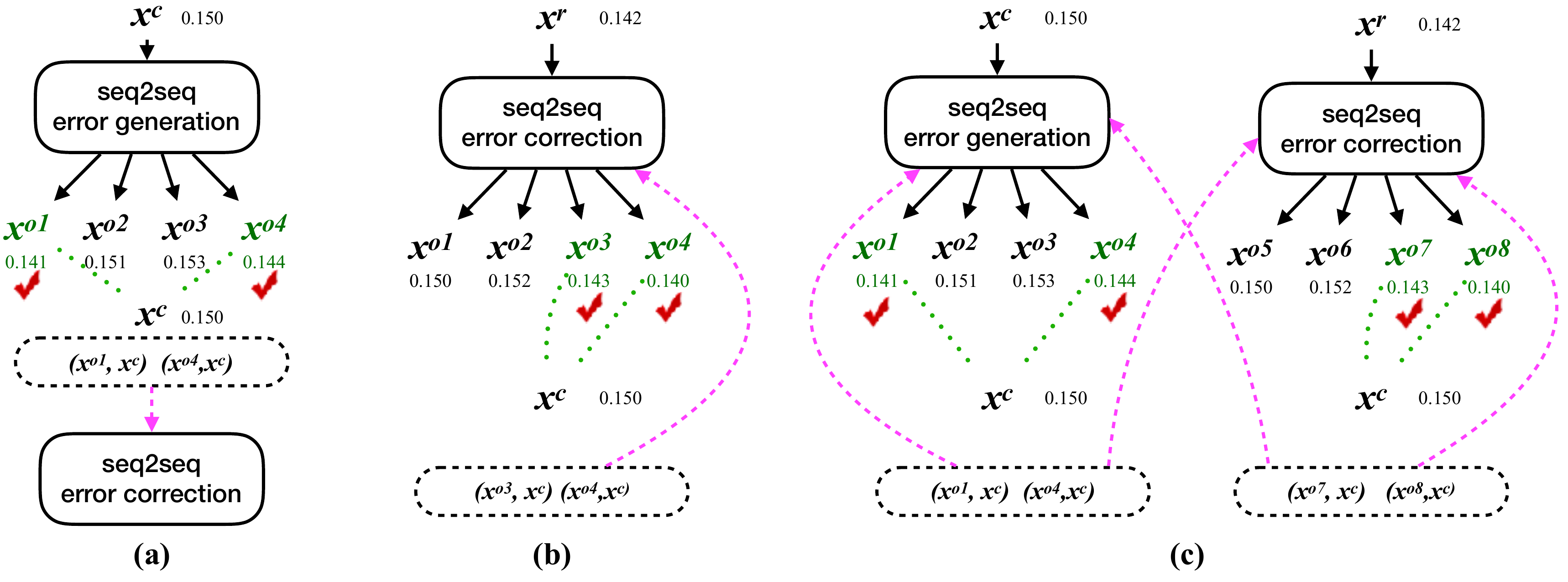}\vspace{-0.4cm}
\caption{Three fluency boost learning strategies: \textbf{(a)} back-boost, \textbf{(b)} self-boost, \textbf{(c)} dual-boost; all of them generate fluency boost sentence pairs (the pairs in the dashed boxes) to help model learning during training. The numbers in this figure are fluency scores of their corresponding sentences.}\label{fig:fbl}
\end{figure}

The core idea of fluency boost learning is to generate fluency boost sentence pairs that satisfy the fluency boost condition during training, as Figure \ref{fig:fb}(a) illustrates, so that these pairs can further help model learning.


In this section, we present three fluency boost learning strategies: back-boost, self-boost, and dual-boost that generate fluency boost sentence pairs in different ways, as illustrated in Figure \ref{fig:fbl}.

\subsection{Back-boost learning}\label{subsec:backboost}

Back-boost learning borrows the idea from back translation \citep{sennrich2016improving} in NMT, referring to training a backward model (we call it error generation model, as opposed to error correction model) that is used to convert a fluent sentence to a less fluent sentence with errors. Since the less fluent sentences are generated by the error generation seq2seq model trained with error-corrected data, they usually do not change the original sentence's meaning; thus, they can be paired with their correct sentences, establishing fluency boost sentence pairs that can be used as training instances for error correction models, as Figure \ref{fig:fbl}(a) shows.

Specifically, we first train a seq2seq error generation model $\boldsymbol{\Theta_{gen}}$ with $\widetilde{\mathcal{S}^*}$ which is identical to $\mathcal{S}^*$ except that the source sentence and the target sentence are interchanged. Then, we use the model $\boldsymbol{\Theta_{gen}}$ to predict $n$-best outputs $\boldsymbol{x^{o_1}, \cdots, x^{o_n}}$ given a correct sentence $\boldsymbol{x^c}$. Given the fluency boost condition, we compare the fluency of each output $\boldsymbol{x^{o_k}}$ (where $1 \le k \le n$) to that of its correct sentence $\boldsymbol{x^c}$. If an output sentence's fluency score is much lower than its correct sentence, we call it \textbf{a disfluency candidate} of $\boldsymbol{x^c}$.

To formalize this process, we first define $\mathcal{Y}_n(\boldsymbol{x};\boldsymbol{\Theta})$ to denote the $n$-best outputs predicted by model $\boldsymbol{\Theta}$ given the input $\boldsymbol{x}$. Then, disfluency candidates of a correct sentence $\boldsymbol{x^c}$ can be derived:

\vspace{-0.2cm}
\begin{equation}\label{eq:dback}
\mathcal{D}_{back}(\boldsymbol{x^c}) = \{ \boldsymbol{x^{o_k}} | \boldsymbol{x^{o_k}} \in \mathcal{Y}_n(\boldsymbol{x_c};\boldsymbol{\Theta}_{gen}) \wedge \frac{f(\boldsymbol{x^{c}})}{f(\boldsymbol{x^{o_k}})} \ge \sigma  \}
\end{equation}

\noindent where $\mathcal{D}_{back}(\boldsymbol{x^c})$ denotes the disfluency candidate set for $\boldsymbol{x^c}$ in back-boost learning. $\sigma$ is a threshold to determine if $\boldsymbol{x^{o_k}}$ is less fluent than $\boldsymbol{x^c}$ and it should be slightly larger\footnote{We set $\sigma=1.05$ since the corrected sentence in our training data improves its corresponding raw sentence about 5\% fluency on average.} than $1.0$, which helps filter out sentence pairs with unnecessary edits (e.g., I like this book. $\to$ I like the book.).

In the subsequent training epochs, the error correction model will not only learn from the original error-corrected sentence pairs ($\boldsymbol{x^r}$,$\boldsymbol{x^c}$), but also learn from fluency boost sentence pairs ($\boldsymbol{x^{o_k}}$,$\boldsymbol{x^c}$) where $\boldsymbol{x^{o_k}}$ is a sample of $\mathcal{D}_{back}(\boldsymbol{x^c}$).

We summarize this process in Algorithm \ref{alg:back-boost} where $\mathcal{S}^*$ is the set of original error-corrected sentence pairs, and $\mathcal{S}$ can be tentatively considered identical to $\mathcal{S}^*$ when there is no additional native data to help model training (see Section \ref{subsec:native}). Note that we constrain the size of $\mathcal{S}_t$ not to exceed $|\mathcal{S}^*|$ (the 7th line in Algorithm \ref{alg:back-boost}) to avoid that too many fluency boost pairs overwhelm the effects of the original error-corrected pairs on model learning.

\begin{algorithm}[t]
\centering
\caption{Back-boost learning\label{alg:back-boost}}
\begin{algorithmic}[1]
\State Train error generation model $\boldsymbol{\Theta_{gen}}$ with $\widetilde{\mathcal{S}^*}$;
\For{each sentence pair $(\boldsymbol{x^r}, \boldsymbol{x^c}) \in \mathcal{S}$}
\State Compute $\mathcal{D}_{back}(\boldsymbol{x^c})$ according to Eq (\ref{eq:dback});
\EndFor
\For{each training epoch $t$}
\State $\mathcal{S'} \gets \emptyset$;
\State Derive a subset $\mathcal{S}_t$ by randomly sampling $|\mathcal{S}^*|$ elements from $\mathcal{S}$;
\For{each $(\boldsymbol{x^r}, \boldsymbol{x^c}) \in \mathcal{S}_t$}
\State Establish a fluency boost pair $(\boldsymbol{x'},\boldsymbol{x^c})$ by randomly sampling $\boldsymbol{x'} \in \mathcal{D}_{back}(\boldsymbol{x^c})$;
\State $\mathcal{S'} \gets \mathcal{S'} \cup \{(\boldsymbol{x'},\boldsymbol{x^c})\}$;
\EndFor
\State Update error correction model $\boldsymbol{\Theta_{crt}}$ with $\mathcal{S}^* \cup \mathcal{S'}$;
\EndFor
\end{algorithmic}
\end{algorithm}

\subsection{Self-boost learning}\label{subsec:selfboost}

In contrast to back-boost learning whose core idea is originally from NMT, self-boost learning is original, which is specially devised for neural GEC. The idea of self-boost learning is illustrated by Figure \ref{fig:fbl}(b) and was already briefly introduced in Section \ref{sec:intro} and Figure \ref{fig:fb}(a). Unlike back-boost learning in which an error generation seq2seq model is trained to generate disfluency candidates, self-boost learning allows the error correction model to generate the candidates by itself. Since the disfluency candidates generated by the error correction seq2seq model trained with error-corrected data rarely change the input sentence's meaning; thus, they can be used to establish fluency boost sentence pairs.

For self-boost learning, given an error corrected pair $(\boldsymbol{x^r},\boldsymbol{x^c})$, an error correction model $\boldsymbol{\Theta_{crt}}$ first predicts $n$-best outputs $\boldsymbol{x^{o_1}}, \cdots, \boldsymbol{x^{o_n}}$ for the raw sentence $\boldsymbol{x^r}$. Among the $n$-best outputs, any output that is not identical to $\boldsymbol{x^c}$ can be considered as an error prediction. Instead of treating the error predictions useless, self-boost learning fully exploits them. Specifically, if an error prediction $\boldsymbol{x^{o_k}}$ is much less fluent than that of its correct sentence $\boldsymbol{x^c}$, it will be added to $\boldsymbol{x^c}$'s disfluency candidate set $\mathcal{D}_{self}(\boldsymbol{x^c})$, as Eq (\ref{eq:self-boost}) shows:

\vspace{-0.5cm}
\begin{equation}\label{eq:self-boost}
\mathcal{D}_{self}(\boldsymbol{x^c}) =  \mathcal{D}_{self}(\boldsymbol{x^c})~\cup~\{ \boldsymbol{x^{o_k}} | \boldsymbol{x^{o_k}} \in \mathcal{Y}_n(\boldsymbol{x_r};\boldsymbol{\Theta}_{crt}) \wedge \frac{f(\boldsymbol{x^{c}})}{f(\boldsymbol{x^{o_k}})} \ge \sigma  \}
\end{equation}
\vspace{-0.4cm}


In contrast to back-boost learning, self-boost generates disfluency candidates from a different perspective -- by editing the raw sentence $\boldsymbol{x^r}$ rather than the correct sentence $\boldsymbol{x^c}$. It is also noteworthy that $\mathcal{D}_{self}(\boldsymbol{x^c})$ is incrementally expanded because the error correction model $\boldsymbol{\Theta_{crt}}$ is dynamically updated,
as shown in Algorithm \ref{alg:self-boost}. 

\begin{algorithm}[h]
\centering
\caption{Self-boost learning\label{alg:self-boost}}
\begin{algorithmic}[1]
\For{each sentence pair $(\boldsymbol{x^r}, \boldsymbol{x^c}) \in \mathcal{S}$}
\State $\mathcal{D}_{self}(\boldsymbol{x^c}) \gets \emptyset$;
\EndFor
\State $\mathcal{S'} \gets \emptyset$
\For{each training epoch $t$}
\State Update error correction model $\boldsymbol{\Theta_{crt}}$ with $\mathcal{S}^* \cup \mathcal{S'}$;
\State $\mathcal{S'} \gets \emptyset$
\State Derive a subset $\mathcal{S}_t$ by randomly sampling $|\mathcal{S}^*|$ elements from $\mathcal{S}$; 
\For{each $(\boldsymbol{x^r}, \boldsymbol{x^c}) \in \mathcal{S}_t$}
\State Update $\mathcal{D}_{self}(\boldsymbol{x^c})$ according to Eq (\ref{eq:self-boost});
\State Establish a fluency boost pair $(\boldsymbol{x'},\boldsymbol{x^c})$ by randomly sampling $\boldsymbol{x'} \in \mathcal{D}_{self}(\boldsymbol{x^c})$;
\State $\mathcal{S'} \gets \mathcal{S'} \cup \{(\boldsymbol{x'},\boldsymbol{x^c})\}$;
\EndFor
\EndFor
\end{algorithmic}
\end{algorithm}

\subsection{Dual-boost learning}\label{subsec:dualboost}

As introduced above, back- and self-boost learning generate disfluency candidates from different perspectives to create more fluency boost sentence pairs to benefit training the error correction model. Intuitively, the more diverse disfluency candidates generated, the more helpful for training an error correction model. Inspired by \cite{he2016dual} and \cite{zhang2018joint}, we propose a dual-boost learning strategy, combining both back- and self-boost's perspectives to generate disfluency candidates. 

As Figure \ref{fig:fbl}(c) shows, disfluency candidates in dual-boost learning are from both the error generation model and the error correction model :

\vspace{-0.4cm}
\begin{equation}\label{eq:dual-boost}
\mathcal{D}_{dual}(\boldsymbol{x^c}) = \mathcal{D}_{dual}(\boldsymbol{x^c}) ~ \cup ~ \{ \boldsymbol{x^{o_k}} | \boldsymbol{x^{o_k}} \in \mathcal{Y}_n(\boldsymbol{x_r};\boldsymbol{\Theta}_{crt}) \cup \mathcal{Y}_n(\boldsymbol{x_c};\boldsymbol{\Theta}_{gen}) \wedge \frac{f(\boldsymbol{x^{c}})}{f(\boldsymbol{x^{o_k}})} \ge \sigma  \}
\end{equation}\vspace{-0.2cm}

Moreover, the error correction model and the error generation model are dual and both of them are dynamically updated, which improves each other: the disfluency candidates produced by error generation model can benefit training the error correction model, while the disfluency candidates created by error correction model can be used as training data for the error generation model. We summarize this learning approach in Algorithm \ref{alg:dual-boost}.

\begin{algorithm}[t]
\centering
\caption{Dual-boost learning\label{alg:dual-boost}}
\begin{algorithmic}[1]
\For{each $(\boldsymbol{x^r}, \boldsymbol{x^c}) \in \mathcal{S}$}
\State $\mathcal{D}_{dual}(\boldsymbol{x^c}) \gets \emptyset$;
\EndFor
\State $\mathcal{S'} \gets \emptyset$; $\mathcal{S''} \gets \emptyset$; 
\For{each training epoch $t$}
\State Update error correction model $\boldsymbol{\Theta_{crt}}$ with $\mathcal{S}^* \cup \mathcal{S'}$;
\State Update error generation model $\boldsymbol{\Theta_{gen}}$ with $\widetilde{\mathcal{S}^*} \cup \mathcal{S''}$;
\State $\mathcal{S'} \gets \emptyset$; $\mathcal{S''} \gets \emptyset$; 
\State Derive a subset $\mathcal{S}_t$ by randomly sampling $|\mathcal{S}^*|$ elements from $\mathcal{S}$; 
\For{each $(\boldsymbol{x^r}, \boldsymbol{x^c}) \in \mathcal{S}_t$}
\State Update $D_{dual}(\boldsymbol{x^c})$ according to Eq (\ref{eq:dual-boost});
\State Establish a fluency boost pair $(\boldsymbol{x'},\boldsymbol{x^c})$ by randomly sampling $\boldsymbol{x'} \in \mathcal{D}_{dual}(\boldsymbol{x^c})$;
\State $\mathcal{S'} \gets \mathcal{S'} \cup \{(\boldsymbol{x'},\boldsymbol{x^c})\}$;
\State Establish a reversed fluency boost pair $(\boldsymbol{x^c},\boldsymbol{x''})$ by randomly sampling $\boldsymbol{x''} \in \mathcal{D}_{dual}(\boldsymbol{x^c})$;
\State $\mathcal{S''} \gets \mathcal{S''} \cup \{(\boldsymbol{x^c},\boldsymbol{x''})\}$;
\EndFor
\EndFor
\end{algorithmic}
\end{algorithm}

\subsection{Fluency boost learning with large-scale native data}\label{subsec:native}


Our proposed fluency boost learning strategies can be easily extended to utilize massive native text data which proved to be useful for GEC.

As discussed in Section \ref{subsec:backboost}, when there is no additional native data, $\mathcal{S}$ in Algorithm \ref{alg:back-boost}--\ref{alg:dual-boost} is identical to $\mathcal{S}^*$. In the case where additional native data is available to help model learning, $\mathcal{S}$ becomes:

\vspace{-0.5cm}
\begin{displaymath}
\mathcal{S}=\mathcal{S}^* \cup \mathcal{C}
\end{displaymath}\vspace{-0.6cm}

\noindent where $\mathcal{C}=\{(\boldsymbol{x^c},\boldsymbol{x^c})\}$ denotes the set of self-copied sentence pairs from native data.

\section{Fluency boost inference}
\subsection{Multi-round error correction}\label{subsec:multiround}

As we discuss in Section \ref{sec:intro}, some sentences with multiple grammatical errors usually cannot be perfectly corrected through normal seq2seq inference which makes only single-round inference. Fortunately, neural GEC is different from NMT: its source and target language are the same. The characteristic allows us to edit a sentence more than once through multi-round model inference, which motivates our fluency boost inference. As Figure \ref{fig:fb}(b) shows, fluency boost inference allows a sentence to be incrementally edited through multi-round seq2seq inference as long as the sentence's fluency can be improved. Specifically, an error correction seq2seq model first takes a raw sentence $\boldsymbol{x^r}$ as an input and outputs a hypothesis $\boldsymbol{x^{o_1}}$. Instead of regarding $\boldsymbol{x^{o_1}}$ as the final prediction, fluency boost inference will then take $\boldsymbol{x^{o_1}}$ as the input to generate the next output $\boldsymbol{x^{o_2}}$. The process will not terminate unless $\boldsymbol{x^{o_t}}$ does not improve $\boldsymbol{x^{o_{t-1}}}$ in terms of fluency. 

\subsection{Round-way error correction}\label{subsec:round-way}
\begin{figure}
\centering
\includegraphics[width=12cm]{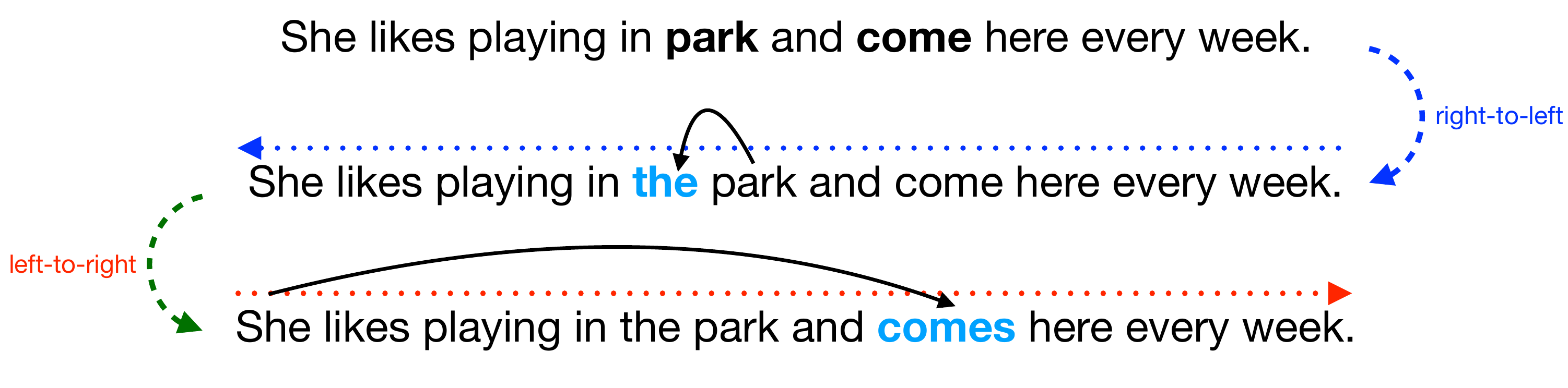}
\caption{Round-way error correction: some types of errors (e.g., articles) are easier to be corrected by a right-to-left seq2seq model, while some (e.g., subject verb agreement) are more likely to be corrected by a left-to-right seq2seq model. Round-way error correction makes left-to-right and right-to-left seq2seq models well complement each other, enabling it to correct more grammatical errors than an individual model.}\label{fig:round}
\end{figure}

Based on the idea of multi-round correction, we further propose an advanced fluency boost inference approach: round-way error correction. Instead of progressively correcting a sentence with the same seq2seq model as introduced in Section \ref{subsec:multiround}, round-way correction corrects a sentence through a right-to-left seq2seq model and a left-to-right seq2seq model successively, as shown in Figure \ref{fig:round}.

The motivation of round-way error correction is straightforward. Decoders with different decoding orders decode word sequences with different contexts, making them have their unique advantages for specific error types. For the example in Figure \ref{fig:round}, the error of a lack of an article (i.e., \textit{park $\to$ the park}) is more likely to be corrected by the right-to-left seq2seq model than the left-to-right one, because whether to add an article depends on the noun \textit{park} that was already seen by the right-to-left model when it made the decision. In contrast, the left-to-right
model might be better at dealing with subject-verb agreement errors (e.g., \textit{come $\to$ comes} in Figure \ref{fig:round}) because the keyword that decides the verb form is its subject \textit{She} which is at the beginning of the sentence.

\section{Experiments}
\subsection{Dataset and evaluation}\label{subsec:data}

\begin{table}[h]
\centering
\begin{tabular}{c|c}
\hline
\textbf{Corpus} & \textbf{\#sent pair} \\ 
\hline
Lang-8          &            1,114,139         \\
CLC             &           1,366,075           \\
NUCLE           &    57,119        \\
Extended Lang-8 &    2,865,639 \\
\hline
\bf Total	&  5,402,972 \\
\hline
\end{tabular}
\caption{Error-corrected training data.}
\label{tab:data_stats}
\end{table}

As previous studies \citep{ji2017nested}, we use the public Lang-8 Corpus \citep{mizumoto2011mining,tajiri2012tense}, Cambridge Learner Corpus (CLC) \citep{nicholls2003cambridge} and NUS Corpus of Learner English (NUCLE) \citep{dahlmeier2013building} as our original error-corrected training data. Table \ref{tab:data_stats} shows the stats of the datasets. In addition, we also collect 2,865,639 non-public error-corrected sentence pairs from Lang-8.com. The native data we use for fluency boost learning is English Wikipedia that contains 61,677,453 sentences.

We use CoNLL-2014 shared task dataset \citep{ng2014conll} and JFLEG \citep{napoles2017jfleg} test set as our evaluation datasets. CoNLL-2014 test set contains 1,312 sentences, while JFLEG test set has 747 sentences. Being consistent with the official evaluation metrics, we use MaxMatch (M$^2$) $F_{0.5}$ \citep{dahlmeier-ng:2012:NAACL-HLT} for CoNLL-2014 and use GLEU \citep{napoles2015ground} for JFLEG evaluation. It is notable that the original annotations for CoNLL-2014 dataset are from 2 human annotators, which are later enriched by \cite{bryant2015far} that contains 10 human expert annotations for each test sentence. We evaluate systems' performance using both annotation settings for the CoNLL dataset. To distinguish between these two annotation settings, we use CoNLL-2014 to denote the original annotations, and CoNLL-10 to denote the 10-human annotations.
As previous studies, we use CoNLL-2013 test set and JFLEG dev set as our development sets for CoNLL-2014 and JFLEG test set respectively.

\subsection{Experimental setting}\label{subsec:setting}

We use 7-layer convolutional seq2seq models\footnote{\url{https://github.com/pytorch/fairseq}} \citep{gehring2017convolutional} as our error correction and error generation model, which have proven to be effective for GEC \citep{chollampatt2018}. As \cite{chollampatt2018}, we set the dimensionality of word embeddings in both encoders and decoders to 500, the hidden size of encoders and decoders to 1,024 and the convolution window width to 3. The vocabularies of the source and target side are the most frequent 30K BPE tokens for each. We train the seq2seq models using Nesterov Accelerated Gradient \citep{sutskever2013importance} optimizer with a momentum
value of 0.99. The initial learning rate is set to 0.25 and it will be reduced by an order of magnitude if the validation perplexity stops
improving. During training, we allow each batch to have at most 3,000 tokens per GPU and set dropout rate to 0.2. We terminate the training process when the learning rate falls below $10^{-4}$. As \cite{chollampatt2018} and \cite{grundkiewicz2018near}, we train 4 models with different random initializations for ensemble decoding.

For fluency boost learning, we adopt dual-boost learning introduced in Section \ref{subsec:dualboost} and use the English Wikipedia data as our native data (Section \ref{subsec:native}). Disfluency candidates are generated from 10-best outputs. For fluency boost inference, we use round-way correction approach introduced in Section \ref{subsec:round-way}. The architecture of the right-to-left seq2seq model in round-way correction is the same with the left-to-right\footnote{In cases other than round-way correction, we use left-to-right seq2seq models as our default error correction models.} one except that they decode sentences in the opposite directions. For single-round inference, we follow \cite{chollampatt2018} to generate 12-best predictions and choose the best sentence after re-ranking with edit operation and language model scores. The language model is the 5-gram language model trained on Common Crawl released by \cite{junczys2016phrase}, which is also used for computing fluency score in Eq (\ref{eq:fluency}).

As most of the systems \citep{sakaguchi2017grammatical,chollampatt2018,grundkiewicz2018near} evaluated on JFLEG that use an additional spell checker to resolve spelling errors, we use a public spell checker\footnote{\url{https://azure.microsoft.com/en-us/services/cognitive-services/spell-check/}} to resolve spelling errors in JFLEG as preprocessing.

\begin{table}[t]
\centering
\small
\begin{tabular}{c|c|c|c}
\hline
\multirow{2}{*}{\textbf{System}} & \textbf{CoNLL-2014} & \textbf{CoNLL-10} & \textbf{JFLEG test} \\
                                 & $F_{0.5}$                   & $F_{0.5}$                 & $GLEU$                \\ \hline
No edit                          & -                   & -                 & 40.54               \\
CAMB14                           & 37.33               & 54.30             & 46.04            \\
CAMB16                           & 39.90               & -                 & 52.05               \\
CAMB17 & 51.08 & - & - \\
CUUI & 36.79 & 51.79 & - \\
VT16 & 47.40 & 62.45 & - \\
AMU14 & 35.01 & 50.17 & - \\
AMU16                            & 49.49               & 66.83             & 51.46               \\
NUS16                            & 44.27               & 60.36             & 50.13               \\
NUS17                            & 53.14               & 69.12             & 56.78               \\
NUS18                            & 54.79               & 70.14                 & 57.47               \\
Nested-RNN-seq2seq                   & 45.15               & -                 & 53.41               \\
Back-CNN-seq2seq & 49.0 &  - & 56.6 \\
Adapted-transformer & 55.8 & - & 59.9 \\
SMT-NMT hybrid                   & 56.25               & -             & 61.50               \\ \hline
\multicolumn{1}{l|}{Base convolutional seq2seq}   &  57.95       &          73.19           &        60.87          \\
\multicolumn{1}{l|}{Base + FB learning}      &    \bf 61.34    &   \bf 76.88 &  61.41    \\
\multicolumn{1}{l|}{Base + FB learning and inference}        &        60.00        & 75.72 & \bf 62.42               \\ \hline
\end{tabular}
\caption{Comparison of GEC systems on CoNLL and JFLEG benchmark datasets.}
\label{tab:gec_result}
\end{table}

\subsection{Experimental results}

We compare our systems\footnote{In this report, we do not present a detailed comparison and analysis for different fluency boost learning and inference methods which can be found in \cite{ge2018fluency}.} to the following well-known GEC systems:

\leftmargini=5mm
\begin{itemize}

\item CAMB14, CAMB16 and CAMB17: GEC systems \citep{felice2014grammatical,yuan2016grammatical,yannakoudakis2017neural} developed by Cambridge University. For CAMB17, we report its best result.
\item CUUI and VT16: the former system \citep{rozovskaya2014illinois} uses a classifier-based approach, which is improved by the latter system \citep{rozovskaya2016grammatical} through combining it with an SMT-based approach.
\item AMU14 and AMU16: SMT-based GEC systems \citep{junczys2014amu,junczys2016phrase} developed by AMU.
\item NUS14, NUS16, NUS17 and NUS18: The first three GEC systems \citep{Susanto2014System,chollampatt2016adapting,chollampatt-ng:2017:BEA} are SMT-based GEC systems that are combined with other techniques (e.g., classifiers). The last one \citep{chollampatt2018} uses convolutional seq2seq models for grammatical error correction.
\item Nested-RNN-seq2seq: a Recurrent Neural Network (RNN) seq2seq model with nested attention \citep{ji2017nested}.
\item Back-CNN-seq2seq: a convolutional seq2seq model \citep{xie2018noising} trained with synthesized data augmented by back translation. Its core idea is somewhat similar to the idea introduced in Section \ref{subsec:backboost} and Section \ref{subsec:native} of this work.
\item Adapted-transformer: a transformer \citep{vaswani2017attention} based GEC system \citep{junczys2018approaching} with techniques adapted from low-resource machine translation.
\item SMT-NMT hybrid: the state-of-the-art GEC system \citep{grundkiewicz2018near} that is based on an SMT-NMT hybrid approach.
\end{itemize}

Table \ref{tab:gec_result} shows the results\footnote{A result marked with ``-'' means that the system's result in the corresponding dataset or setting is not reported by the original papers or other literature and that the system outputs are not publicly available.} of GEC systems on CoNLL and JFLEG dataset. Our base convolutional seq2seq model outperforms most of previous GEC systems owing to the larger size of training data we use. Fluency boost learning further improves the base convolutional seq2seq model. It achieves 61.34 in CoNLL-2014, 76.88 $F_{0.5}$ score in CoNLL-10 benchmarks, and 61.41 GLEU score on JFLEG test set. When we further add fluency boost inference, the system's performance on JFLEG test set is improved to 62.42 GLEU score, while its $F_{0.5}$ scores on CoNLL benchmarks drop.

\begin{table}[t]
\small
\scalebox{0.9}
{
\begin{tabular}{c|ccc|ccc|c|c}
\hline
\multirow{2}{*}{\bf System}                              & \multicolumn{3}{c|}{CoNLL-2014}                   & \multicolumn{3}{c|}{CoNLL-10}                     & CoNLL-10 (SvH) & JFLEG          \\ 
                                                     & $P$              & $R$              & $F_{0.5}$              & $P$              & $R$              & $F_{0.5}$              & $F_{0.5}$              & $GLEU$           \\ \hline
NUS17                                                & 62.74          & 32.96          & 53.14          & 80.04              & 44.71              & 69.12          & 68.29          & 56.78          \\
NUS18                                                & 65.49          & 33.14          & 54.79          & 81.05              & 45.60              & 70.14              & 69.30              & 57.47          \\
Adapted-transformer &  61.9 & \bf 40.2 & 55.8 & - & - & - & - & 59.9 \\
SMT-NMT hybrid                                       & 66.77          & 34.49          & 56.25          & -              & -              & -          & 72.04             & 61.50          \\ \hline
\multicolumn{1}{l|}{Base convoluation seq2seq}        & 72.52          & 32.13          & 57.95          & 86.65          & 45.14          & 73.19          & 72.28          & 60.87          \\
\multicolumn{1}{l|}{Base + FB learning}               & \textbf{74.12} & 36.30          & \textbf{61.34} & \textbf{88.56} & 50.31          & \textbf{76.88} & \color{red}{\textbf{75.93}} & 61.41          \\
\multicolumn{1}{l|}{Base + FB learning and inference} & 68.45          & 40.18 & 60.00          & 84.71          & \textbf{53.15} & 75.72          & \color{red}{74.84}          & \color{red}{\textbf{62.42}} \\ \hline
Human performance                                    & -              & -              & -              & -              & -              & -              & 72.58          & 62.37         \\ \hline 
\end{tabular}
}
\caption{Evaluation result analysis for top-performing GEC systems on CoNLL and JFLEG datasets. The results marked with red font exceed the human-level performance.}
\label{tab:result_analysis}
\end{table}

We look into the results in Table \ref{tab:result_analysis}. Fluency boost learning improves the base convolutional seq2seq model in terms of all aspects (i.e., precision, recall, $F_{0.5}$ and GLEU), demonstrating fluency boost learning is actually helpful for training a seq2seq model for GEC. Adding fluency boost inference improves recall (from 36.30 to 40.18 on CoNLL-2014 and from 50.31 to 53.15 on CoNLL-10) at the expense of a drop of precision (from 74.12 to 68.45 on CoNLL-2014 and from 88.56 to 84.71 on CoNLL-10). Since $F_{0.5}$ weighs precision twice as recall, adding fluency boost inference leads to a drop of $F_{0.5}$ on the CoNLL dataset. In contrast, for JFLEG, fluency boost inference improves GLEU score from 61.41 to 62.42, demonstrating its effectiveness for improving sentences' fluency.

We compare our systems to human performance on CoNLL-10 and JFLEG benchmarks. For CoNLL-10, we follow the evaluation setting in \cite{bryant2015far} and \cite{chollampatt-ng:2017:BEA} to fairly compare systems' performance to human's, which is marked with (SvH) in Table \ref{tab:result_analysis}. Among our systems, the system with fluency boost learning and inference outperforms human's performance on both CoNLL and JFLEG dataset, while the system with only fluency boost learning achieves higher $F_{0.5}$ scores on CoNLL dataset.

\begin{table}[t]
\centering
\small
\begin{tabular}{c|c|c}
\hline
\bf Error type & \bf Base convolutional seq2seq & \bf Base + fluency boost learning \\ \hline
ArtOrDet   & 26.00                      & \bf 28.26              \\
Mec        & 25.45                      & \bf 25.54              \\
Nn         & 46.10                      & \bf 53.99              \\
Npos       & 20.00                      & \bf 24.00              \\
Pform      & \bf 17.54                      & 15.79              \\
Pref       & 4.69                       & \bf 7.04               \\
Prep       & 23.38                      & \bf  28.51              \\
Rloc       & 9.54                       & 9.54               \\
Sfrag      & 0                          & \bf 7.14               \\
Smod       & 0                          & 0                  \\
Spar       & 8.00                       & \bf 12.00              \\
Srun       & 0                          & 0                  \\
Ssub       & 10.14                      & \bf 14.49              \\
SVA        & 34.74                      & \bf 42.11              \\
Trans      & 5.63                       & \bf 8.45               \\
Um         & 2.04                       & 2.04               \\
V0         & 23.21                      & \bf 26.79              \\
Vform      & 34.81                      & \bf 42.78              \\
Vm         & 11.69                      & 11.69              \\
Vt         & 14.36                      & \bf 19.70              \\
Wa         & 0                          & 0                  \\
Wci        & 7.50                       & \bf 9.15               \\
Wform      & 43.28                      & \bf 47.01              \\
WOadv      & 5.88                       & \bf 23.53              \\
WOinc      & 1.45                       & \bf 4.35               \\
Wtone      & 8.70                       & \bf 17.39              \\
Others     & 0                          & \bf 1.22              \\ \hline
\end{tabular}
\caption{A comparison of recall of the convolutional seq2seq model with/without fluency boost learning for each error type in CoNLL-2014 dataset.}
\label{tab:boost_errortype}
\end{table}

We further study the effectiveness of fluency boost learning and inference for different error types. Table \ref{tab:boost_errortype} shows the recall of base convolutional seq2seq model and the model trained with fluency boost learning for each error type\footnote{The definitions of error types in Table \ref{tab:boost_errortype} can be found in \cite{ng2014conll}.} in CoNLL-2014 dataset (original annotation setting). One can see that fluency boost learning improves recall for most error types, demonstrating that fluency boost learning approach can generate sentences with diverse errors to help training.

To better understand the effectiveness of fluency boost inference (i.e., round-way error correction), we show in Table \ref{tab:errortype} the recall of each error type of the left-to-right and the right-to-left seq2seq in CoNLL-2014 dataset (original annotation setting). Note that to clearly see pros and cons of the left-to-right and right-to-left model, here we do not re-rank their n-best results using edit operations and the language model; instead, we directly use their 1-best generated sentence as their prediction.

According to Table \ref{tab:errortype}, the right-to-left model does better in the error types like ArtOrDet, while the left-to-right model is better at correcting the errors like SVA, which is consistent with our motivation in Section \ref{subsec:round-way}. When we use round-way correction, the errors that are not corrected by the right-to-left model are likely to be corrected by the left-to-right one, which is reflected by the recall improvement of most error types, as shown in Table \ref{tab:errortype}.

\begin{table}[t]
\small
\centering
\begin{tabular}{c|c|c|c}
\hline
\textbf{Error type} & \textbf{Right-to-Left} & \textbf{Left-to-Right} & \textbf{Round-way (R2L $\to$ L2R)} \\ \hline
ArtOrDet            & \textbf{25.70}         & 22.31                  & 30.36                                                          \\
Mec                 & 16.27                  & 16.52                  & 20.40                                                          \\
Nn                  & 32.13                  & \textbf{38.03}         & 41.31                                                          \\
Npos                & \textbf{16.00}         & 12.00                  & 16.00                                                          \\
Pform               & \textbf{17.54}         & 14.04                  & 19.30                                                          \\
Pref                & 2.35                   & 2.35                   & 3.76                                                           \\
Prep                & 14.88                  & \textbf{17.40}         & 21.81                                                          \\
Rloc                & 7.25                   & 6.87                   & 9.92                                                           \\
Sfrag               & 0                      & 0                      & 0                                                              \\
Smod                & 0                      & 0                      & 0                                                              \\
Spar                & 4.00                   & \textbf{12.00}         & 8.00                                                           \\
Srun                & 0                      & 0                      & 0                                                              \\
Ssub                & \textbf{7.25}          & 5.80                   & 10.14                                                          \\
SVA                 & 30.85                  & \textbf{36.84}         & 39.47                                                          \\
Trans               & \textbf{7.04}          & 4.93                   & 7.04                                                           \\
Um                  & \textbf{2.04}          & 0                      & 2.04                                                           \\
V0                  & \textbf{21.43}         & 17.86                  & 28.57                                                          \\
Vform               & 25.14                  & \textbf{31.67}         & 33.52                                                          \\
Vm                  & \textbf{7.79}          & 6.49                   & 9.09                                                           \\
Vt                  & \textbf{13.37}         & 11.33                  & 14.36                                                          \\
Wa                  & 0                      & 0                      & 0                                                              \\
Wci                 & \textbf{5.50}          & 4.68                   & 6.67                                                           \\
Wform               & 35.34                  & 37.59                  & 41.04                                                          \\
WOadv               & 8.82                   & \textbf{14.71}         & 17.65                                                          \\
WOinc               & 2.90                   & 2.90                   & 4.35                                                           \\
Wtone               & \textbf{8.70}          & 4.35                   & 8.70                                                           \\
Others              & 1.22                   & 1.22                   & 1.22                                                          \\ \hline
\end{tabular}
\caption{The left-to-right and right-to-left seq2seq model's recall of each error type in CoNLL-2014.}\vspace{-0.1cm}
\label{tab:errortype}
\end{table}

\section{Related work}

Most of advanced GEC systems are classifier-based \citep{chodorow2007detection,de2008classifier,han2010using,leacock2010automated,tetreault2010using,dale2011helping}
or MT-based
\citep{brockett2006correcting,dahlmeier2011correcting,dahlmeier2012beam,yoshimoto2013naist,yuan2013constrained,behera2013automated}. For example, top-performing systems \citep{felice2014grammatical,rozovskaya2014illinois,junczys2014amu} in CoNLL-2014 shared task \citep{ng2014conll} use either of the methods. Recently, many novel approaches
\citep{Susanto2014System,chollampatt2016neural,chollampatt2016adapting,rozovskaya2016grammatical,junczys2016phrase,mizumoto2016discriminative,Yuan2016Candidate,Hoang2016Exploiting,yannakoudakis2017neural} have been proposed for GEC. Among them, seq2seq models \citep{yuan2016grammatical,xie2016neural,ji2017nested,sakaguchi2017grammatical,schmaltz-EtAl:2017:EMNLP2017,chollampatt2018,junczys2018approaching} have caught much attention. 
Unlike the models trained only with original error-corrected data, we propose a novel fluency boost learning mechanism for dynamic data augmentation along with training for GEC, despite some related studies that explore artificial error generation for GEC \citep{brockett2006correcting,foster2009generrate,rozovskaya2010training,rozovskaya2011algorithm,Rozovskaya2012The,felice-yuan:2014:SRW,xie2016neural,rei2017artificial,xie2018noising}. Moreover, we propose fluency boost inference which allows the model to repeatedly edit a sentence as long as the sentence's fluency can be improved. To the best of our knowledge, it is the first to conduct multi-round seq2seq inference for GEC, while similar ideas have been proposed for NMT \citep{DXiaTWLQYL17}.


In addition to the studies on GEC, there is also much research on grammatical error detection \citep{leacock2010automated,rei-yannakoudakis:2016:P16-1,kaneko2017grammatical} and GEC evaluation \citep{tetreault2010rethinking,madnani2011they,dahlmeier2012better,napoles2015ground,sakaguchi2016reassessing,napoles2016there,bryant2017automatic,asano2017reference,choshen2018inherent}. 
We do not introduce them in detail because they are not much related to this work's contributions.

\section{Conclusion}

We present a state-of-the-art convolutional seq2seq model based GEC system that uses a novel fluency boost learning and inference mechanism. Fluency boost learning fully exploits both error-corrected data and native data by generating diverse error-corrected sentence pairs during training, which benefits model learning and improves the performance over the base seq2seq model, while fluency boost inference utilizes the characteristic of GEC to progressively improve a sentence's fluency through round-way correction. The powerful learning and inference mechanism enables our system to achieve state-of-the-art results and reach human-level performance on both CoNLL-2014 and JFLEG benchmark datasets.
\bibliography{gec}
\bibliographystyle{iclr2017_conference}

\end{document}